\documentclass[lettersize,journal]{IEEEtran}
\usepackage{amsmath,amsfonts}
\usepackage{algorithmic}
\usepackage{algorithm}
\usepackage{array}
\usepackage[caption=false,font=normalsize,labelfont=sf,textfont=sf]{subfig}
\usepackage{textcomp}
\usepackage{stfloats}
\usepackage{url}
\usepackage{verbatim}
\usepackage{graphicx}
\usepackage{booktabs}
\usepackage{multirow}
\usepackage{amssymb}
\usepackage{cite}
\begin{document}

\title{FusionFM: All-in-One Multi-Modal Image Fusion \\
with Flow Matching}

\author{Huayi Zhu, 
		Xiu Shu, 
		Youqiang Xiong, 
		Qiao Liu,
        Rui Chen,~\IEEEmembership{Member,~IEEE,} 
        Di Yuan,~\IEEEmembership{Member,~IEEE,} \\
        Xiaojun Chang,~\IEEEmembership{Senior Member,~IEEE,} 
        and Zhenyu He,~\IEEEmembership{Senior Member,~IEEE} 
        
\thanks{H. Zhu, Y. Xiong, R. Chen and D. Yuan (Corresponding author),  are with Guangzhou Institute of Technology, Xidian University, Guangzhou, 510555 China (e-mail:24181214158@stu.xidian.edu.cn, 24181214212@stu.xidian.edu.cn, rchen@xidian.edu.cn, dyuanhit@gmail.com).}
\thanks{X. Shu is with the School of Computer Science and Cyber Engineering, Guangzhou University, Guangzhou, 510006 China (e-mail:shuxiu@gzhu.edu.cn).}
\thanks{Q. Liu is with the National Center for Applied Mathematics, Chongqing Normal University, Chongqing, 401331 China(e-mail:liuqiao.hit@gmail.com).}
\thanks{X. Chang is with the Australian Artificial Intelligence Institute, Faculty of Engineering and Information Technology, University of Technology Sydney, 2007 Australia (e-mail: xiaojun.chang@uts.edu.au).}
\thanks{Z. He is with the School of Computer Science and Technology, Harbin Institute of Technology, Shenzhen, 518055 China (e-mail:zhenyuhe@hit.edu.cn).}
}
\markboth{Journal of \LaTeX\ Class Files,~Vol.~14, No.~8, August~2021}%
{Shell \MakeLowercase{\textit{et al.}}: A Sample Article Using IEEEtran.cls for IEEE Journals}



\maketitle

\begin{abstract}
Current multi-modal image fusion methods typically rely on task-specific models, leading to high training costs and limited scalability. While generative methods provide a unified modeling perspective, they often suffer from slow inference due to the complex sampling trajectories from noise to image. To address this, we formulate image fusion as a direct probabilistic transport from source modalities to the fused image distribution, leveraging the flow matching paradigm to improve sampling efficiency and structural consistency. To mitigate the lack of high-quality fused images for supervision, we collect fusion results from multiple state-of-the-art models as priors, and employ a task-aware selection function to select the most reliable pseudo-labels for each task. We further introduce a \textit{Fusion Refiner} module that employs a divide-and-conquer strategy to systematically identify, decompose, and enhance degraded components in selected pseudo-labels. For multi-task scenarios, we integrate \textit{elastic weight consolidation} and \textit{experience replay} mechanisms to preserve cross-task performance and enhance continual learning ability from both parameter stability and memory retention perspectives. Our approach achieves competitive performance across diverse fusion tasks, while significantly improving sampling efficiency and maintaining a lightweight model design. The code will be available at: \url{https://github.com/Ist-Zhy/FusionFM}.
\end{abstract}


\section{Introduction}
Multi-modal image fusion (MMIF) plays a crucial role in security surveillance, medical imaging, and related fields \cite{Survey1, Survey2}. By integrating information from different modalities, MMIF enables more comprehensive and robust scene perception \cite{Survey3}. Common MMIF tasks include infrared and visible image fusion (IVF), medical image fusion (MIF), multi-exposure image fusion (MEF), and multi-focus image fusion (MFF).

\begin{figure}[ht]
\centering
\includegraphics[width=\columnwidth]{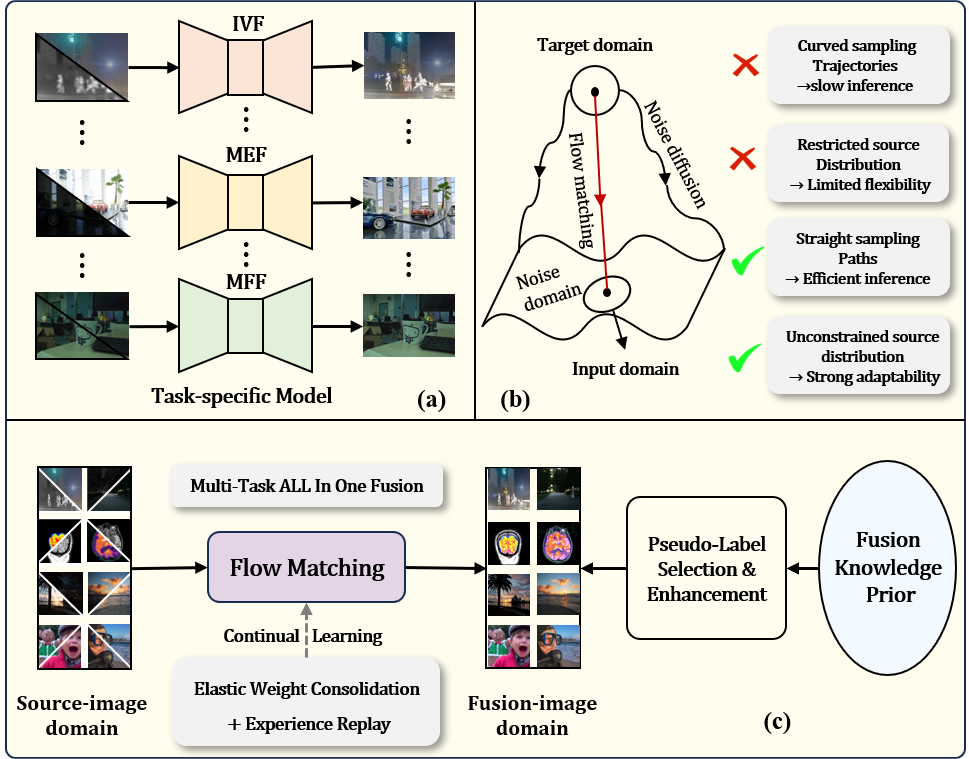} 
\caption{
Motivation of our work. (a) Existing methods rely on “one-task-one-model” and suffer from limited generalization in joint training. (b) Compares traditional diffusion sampling with the efficient and adaptable straight sampling paths of flow matching. (c) Proposed FusionFM Framework.}
\label{Motivation}
\end{figure}

Despite recent advances in the field, MMIF still faces several critical challenges. First, the increasing diversity of fusion tasks and the growing complexity of deployment have become significant obstacles. Most existing approaches adopt a task-specific modeling paradigm, where a separate model must be trained for each individual task \cite{SwinFusion, LFDT, S4Fusion_TIP}. This one-task-one-model strategy incurs high training and deployment costs, making it difficult to adapt to the dynamic and responsive requirements of real-world systems \cite{GIFNet, UMCFuse_TIP}. Secondly, challenges related to joint training and catastrophic forgetting remain pressing \cite{U2Fusion,MAFS_TIP}. While joint training or multi-task learning offers a potential solution to reduce the number of required models, it often leads to catastrophic forgetting \cite{PSLPT,URFusion_TIP}. Moreover, data imbalance across different image fusion tasks exacerbates this issue. For instance, medical image datasets are typically much smaller than those in other domains. Under simple joint training, the model tends to overfit tasks with abundant data while neglecting those with limited samples. Additionally, the computational cost of large-scale joint training is prohibitively high, limiting scalability and practical deployment. Finally, in terms of image quality, state-of-the-art (SOTA) discriminative MMIF models achieve impressive overall performance but still lack fine-grained high-frequency details \cite{EMMA, VDMUFusion}. Recent generative approaches, such as diffusion models, attempt to overcome this limitation by modeling MMIF as an image-conditioned iterative denoising process or by leveraging their powerful feature extraction capabilities \cite{DDFM, Dif-Fusion, Text-DiFuse, FusionINV_TIP}. Although these methods can generate complete fused images, they suffer from extremely slow inference due to the need to integrate along highly curved trajectories of ordinary differential equations (ODEs), which severely limits their applicability in real-time scenarios.

To this end, we introduce Flow Matching (FM) \cite{flow, flow1} as an efficient generative alternative. FM enables a more direct and efficient inference process by explicitly modeling the probabilistic flow from the source distribution to the target distribution, thereby significantly reducing inference time \cite{depthfm, flow2}. In contrast to diffusion models, FM does not constrain the source distribution to Gaussian noise, allowing it to more closely align with the structure of the target data. This flexibility makes FM particularly well-suited for fusion tasks that exhibit strong structural correlations across modalities. We hypothesize that FM offers a more natural and computationally efficient modeling paradigm for MMIF.

While FM offers significant advantages for image fusion, its practical deployment faces two critical challenges. First, real-world scenarios often lack high-quality ground-truth fused images, making it difficult to define target distributions for effective flow learning. Second, developing unified models across multiple fusion tasks requires addressing catastrophic forgetting while enabling efficient knowledge transfer without repeatedly accessing previous task data. To tackle these challenges, we propose FusionFM, integrating external knowledge from pre-trained models with parameter-efficient continual learning. Our approach employs a two-stage pseudo-truth generation strategy: we first leverage outputs from multiple SOTA models as priors, applying task-specific selection to identify reliable pseudo-labels for each task. Inspired by Fusion Boost \cite{Fusionbooster}, we then refine these labels to better align with the data manifold. For continual learning, we combine elastic weight consolidation (EWC) and experience replay (ER), using parameter regularization and strategic memory rehearsal to prevent catastrophic forgetting. FusionFM demonstrates competitive performance across representative fusion tasks. Our main contributions are:
\begin{itemize}
\item We introduce FusionFM, a versatile and efficient all in one MMIF model. By formulating image fusion as a direct distribution transfer from the source image domain to the target fused image domain, FusionFM employs Flow Matching to learn this deterministic flow, enabling fast and high-quality image fusion.
\item We design a two-stage pseudo-truth generation strategy. First, an existing fusion method is used as a soft teacher, combined with a task-aware evaluation metric to filter task-relevant pseudo-labels. Second, inspired by the Fusion Booster concept, we further refine the preliminary pseudo-truth to produce high-fidelity pseudo labels.
\item We propose a novel integration of EWC and ER tailored for MMIF, where EWC preserves modality-specific knowledge while ER maintains cross-modal correlations through strategic sample selection, addressing the unique challenges of forgetting in fusion tasks.
\end{itemize}

\section{Related Work}
\paragraph{Multi-Modal Image Fusion}
Vision-oriented MMIF usually focus on integrating complementary information from multiple modalities and improving visual fidelity. These methods usually rely on complex network architectures and carefully designed loss functions.\cite{Survey4, HoLoCo, B}. In addition, most existing approaches are task-specific, requiring retraining or fine-tuning for new modality combinations, which hinders their generalization and deployment efficiency in multi-task scenarios. To improve generalization, universal fusion frameworks have been proposed \cite{PSLPT}. For example, GIFNet \cite{GIFNet} uses pixel-wise supervision of low-level vision tasks for shared feature learning. TC-MoA \cite{TC-MoA} uses a mixture-of-experts architecture with task-specific adapters to guide a shared backbone across diverse tasks. Although these approaches improve task generality, they still struggle with constantly expanding task sequences and catastrophic forgetting. Meanwhile, recent work has incorporated additional priors and modalities, including joint registration \cite{C2RF}, semantic guidance \cite{Text-if, SeAFusion}, and large language models \cite{FILM}, further extending the scope of MMIF.
\paragraph{Diffusion Models for MMIF}
Owing to their strong generative capacity, diffusion models have recently been widely applied to image processing and extended to MMIF \cite{DRMF, DDFM}. However, their application faces two main challenges. First, the lack of ground truth in image fusion complicates supervised learning. In the absence of real labels, most methods rely on pseudo-labels generated by individual fusion models \cite{MMAIF}, which often introduce bias and limit model generalization. To mitigate this, Diff-IF \cite{Diff-IF} aggregates outputs from multiple models as priors and applies a selection function to produce more reliable pseudo-labels. Second, diffusion models typically incur high computational and inference costs. Although latent diffusion has been explored to reduce overhead, conventional VAE-based latent spaces often fail to capture cross-modal discrepancies in general-purpose fusion tasks, leading to the loss of high-frequency details \cite{OmniFuse, MMAIF, LFDT}. In contrast, Flow Matching \cite{flow1, depthfm} offers a more direct and efficient alternative by modeling continuous vector fields between source and target distributions. This approach significantly improves sampling and training efficiency while preserving fusion quality.
\paragraph{Incremental Learning in MMIF}
Incremental learning aims to enable a model to continuously learn new tasks while retaining its memory of existing tasks to avoid catastrophic forgetting \cite{ICL}. This capability has been widely used in fields such as image recognition and natural language processing \cite{ICL1}. For MMIF tasks, U2Fusion \cite{U2Fusion} adopts an elastic weight consolidation strategy to alleviate the forgetting problem by regularizing key parameters. However, a single EWC approach often lacks sufficient flexibility when faced with large-scale and diverse tasks. To this end, we introduced experience replay based on EWC, combining parameter constraints with sample reproduction to effectively improve the model's continuous learning ability and generalization performance.

\begin{figure}[ht]
\centering
\includegraphics[width=\columnwidth]{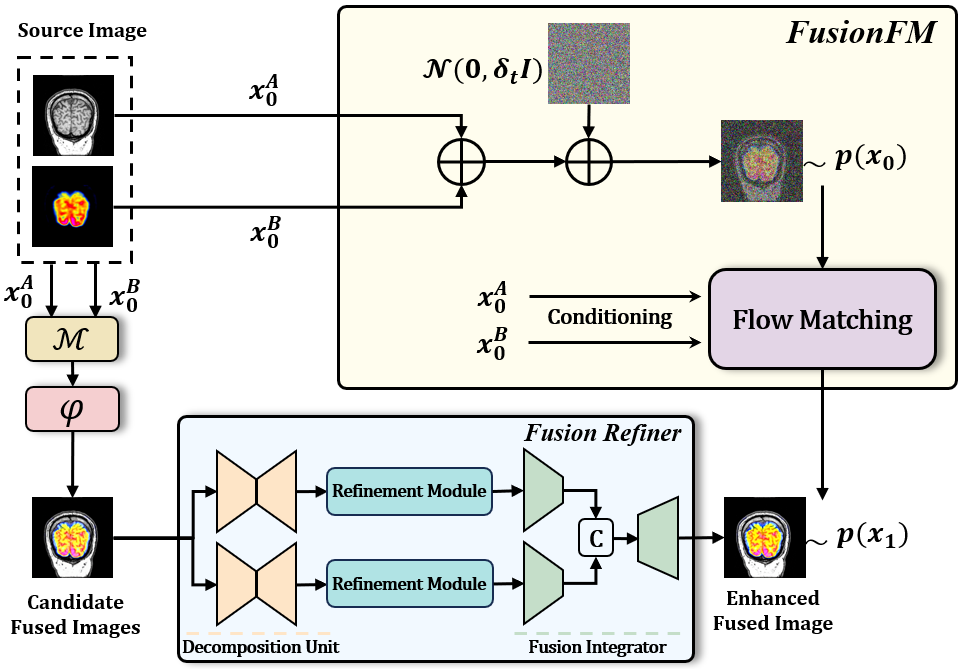} 
\caption{
Overview of our training pipeline. The latent input $x_0$ is formed by adding two modality-specific source images $x_0^A$ and $x_0^B$. A set of fusion candidates is generated by multiple pretrained fusion models $\mathcal{M}$ and scored by the task-aware selector $\varphi$. The top-ranked result is further refined by the Fusion Refiner network to produce the pseudo ground truth $x_1$. Flow Matching is trained to learn the vector field from $x_0$ to $x_1$, conditioned on $(x_0^A, x_0^B)$.
}
\label{o}
\end{figure}

\section{The Proposed Method}
\subsection{Flow Matching-based Image Fusion Model}
To achieve high-quality MMIF, we adopt a FM-based generative model as the core of our framework. FM is a recently proposed generative paradigm that learns a continuous deterministic mapping from a simple prior distribution to a complex data distribution. Unlike diffusion models that rely on iterative denoising, FM models the transformation as a vector field, enabling faster and more stable sampling.
\paragraph{Background: Flow Matching}  
FM is a generative modeling framework that learns a vector field conditioned on fixed probabilistic paths, smoothly transforming samples from a source to a target distribution. Let $\mathbb{R}^d$ denote the data space, and $x$ represent a data point. FM models a time-dependent vector field $u_t(x): [0,1] \times \mathbb{R}^d \rightarrow \mathbb{R}^d$, which governs the evolution of the following ODE: $d{x} = u_t(x)dt$. Let $\phi_t(x)$ be its solution with initial condition $\phi_0(x) = x$. In generative modeling, the goal is to transform entire data distributions rather than individual sample trajectories. Let $p_t: [0,1] \times \mathbb{R}^d \rightarrow \mathbb{R}_{>0}$ denote the time-dependent density function, satisfying $\int p_t(x)dx = 1$. According to the continuity constraint, the pushforward function $p_t = [\phi_t]_{_{\#}}(p_0)$ describes how the initial distribution $p_0$ is transported to any intermediate time $t$ along the vector field $u_t$.

Lipman et al. \cite{flow} proposed a conditional flow matching objective to train neural networks to approximate the conditional vector field $u_t(x | x_1)$, where $x_1$ is sampled from the data distribution and $x$ from the intermediate path $p_t(x | x_1)$. The loss is defined as:
\begin{equation}
\mathcal{L}_{FM}(\theta) = \mathbb{E}_{t, p_{\text{data}}({x}_1), {x} \sim p_t({x} | {x}_1)}  \left\| {v}_\theta(t, {x})-{u}_t({x} | {x}_1) \right\|^2.
\end{equation}
where ${v}_\theta(t, {x})$ is a neural network–parameterized vector field. The objective minimizes its deviation from the true vector field ${u}_t({x} | {x}_1)$. In practice, a common design is to define the path from the source point ${x}_0$ to the target ${x}_1$ via linear interpolation. Assuming the source ${x}_0 = \boldsymbol{\epsilon} \sim \mathcal{N}(\mathbf{0}, \mathbf{I})$, the intermediate point at time $t$ is:
\begin{equation}
{x}_t = t {x}_1 + (1 - t) \boldsymbol{\epsilon},
\end{equation}
with the corresponding analytical vector field expressed as:
\begin{equation}
{u}_t({x} | {x}_1) = \frac{{x}_1 - {x}}{1 - t}.
\end{equation}

This formulation ensures a smooth transition from the noise vector $\boldsymbol{\epsilon}$ to the target sample ${x}_1$, while satisfying the continuity condition of the probability density path.

\paragraph{Direct Transport between Source and Fusion Image}
In the context of MMIF, we reformulate FM not as a transformation from Gaussian noise to a target fused image, but rather as a direct distribution transport from source image features to the fused image features. Unlike diffusion models, which define a conditional distribution $p({x}_1 | \boldsymbol{\epsilon}; {x}_0)$ mapping noise $\boldsymbol{\epsilon}$ to ${x}_1$, we instead model $p({x}_1 | {x}_0)$ to capture the intrinsic transport between source features ${x}_0$ (composed of ${x}_0^A$ and ${x}_0^B$) and the fused target ${x}_1$.

Here, $\{ {x}_0^A, {x}_0^B, {x}_1\}$ represent modality-specific source images and their corresponding fusion. We define the FM source as ${x}_0 = {x}_0^A + {x}_0^B$. As shown in Table~\ref{w}, Average source coupling results in a far shorter transport path to the fusion ground truth compared to random noise coupling across all evaluated image fusion tasks. Unlike noise-based trajectories, this direct transport path is not only shorter but also semantically richer, facilitating more stable and efficient training. Specifically, we define the intermediate interpolated sample between $\mathbf{x}_0$ and ${x}_1$ at time $t \in [0,1]$ as:
\begin{equation}
{x}_t \sim p_t({x} | ({x}_0, {x}_1)) = \mathcal{N}({x} | t{x}_1 + (1 - t){x}_0, \sigma_{\min}^2 \mathbf{I}),
\end{equation}
where $\sigma_{\min}^2$ is a small regularization variance to ensure smooth interpolation and avoid degeneracy. The corresponding vector field that governs the transport from ${x}_0$  to the marginal distribution $p_t({x} \mid ({x}_0, {x}_1))$ is defined as:
\begin{equation}
{u}_t({x} | ({x}_0, {x}_1)) = {x}_1 - {x}_0.
\end{equation}

\begin{table}[t]
\centering
\caption{Average (Avg) source coupling is better than the
Random Coupling. $W$ = Earth Mover’s Distance.}
\resizebox{0.8\columnwidth}{!}{
\begin{tabular}{ccc}
\toprule
Coupling Method & Avg $W_1$($\downarrow$) & Avg $W_2$($\downarrow$) \\
\midrule
Random Coupling & 0.7094 & 0.7663 \\
Average Coupling & \textbf{0.0632} & \textbf{0.0078} \\
\bottomrule
\end{tabular}
}
\label{w}
\end{table}

Although ${x}_0$ and ${x}_1$ may lie on different manifolds due to modality heterogeneity, they form a paired source-fusion instance that implicitly satisfies the conditions for optimal transport. This design effectively frames the dynamic optimal transport problem within the FM framework for image-to-image translation, enabling more stable and efficient convergence. The loss thus takes the form:
\begin{align}
\mathcal{L}(\theta) = \mathbb{E}_{t \sim \mathcal{U}[0,1], ({x}_0^A, {x}_0^B, {x}_1) \sim \mathcal{D}^{GT}} 
&\big[\, \big\| {v}_\theta(t, {x}_t; {x}_0^A, {x}_0^B) \notag \\
&- ({x}_1 - {x}_0) \big\|_1 \,\big],
\end{align}
where $\mathcal{D}^{GT}$ denotes the dataset of paired source and fused images. The vector field is conditioned on both modality inputs via ${v}_\theta(t, {x}_t; {x}_0^A, {x}_0^B)$, rather than solely on ${x}_t$, providing modality-specific guidance for transport learning.

The vector field \(v_\theta(t, x_t; x_0^A, x_0^B)\) in our flow matching framework is parameterized by a \textbf{U-Net} architecture, chosen for its strong multi-scale feature extraction and reconstruction capabilities. The network consists of four downsampling and four upsampling stages, with a \textbf{base channel size of 64} and a \textbf{channel multiplier configuration of \(\mathbf{[1,1,1,1]}\)} indicating the number of residual blocks per stage. The input to the network includes the interpolated sample \(x_t\) concatenated with the source modality features \(x_0^A\) and \(x_0^B\). The output is a vector field prediction that matches the dimensionality of the input features, guiding the transformation from the source domain to the fused image domain.

\subsection{Prior-Guided Pseudo Ground-Truth Generation}
To address the lack of reliable ground truth in supervised image fusion, we propose a prior-guided two-stage pseudo-label generation framework inspired by the concept of knowledge distillation \cite{Diff-IF, Fusionbooster}.

\paragraph{Generation of Candidate Fused Images}
Given a pair of input images $(I_A, I_B)$, we construct a candidate set of fused images by leveraging a pool of SOTA fusion models, denoted as $\mathcal{Q} = {\mathcal{F}_1, \mathcal{F}_2, \dots, \mathcal{F}_Q}$. Each model independently fuses the input pair, yielding the candidate set $\mathcal{S}$:
\begin{equation}
\mathcal{S} = \left\{ I_f^{(1)}, \dots, I_f^{(Q)} \right\}, \text{where}  \ \ I_f^{(q)} = \mathcal{F}_q(I_A, I_B).
\end{equation}

We then apply a task-aware selection function $\varphi$ that ranks the candidates based on evaluation metrics $\text{Score}(\cdot; \tau)$ tailored to the specific fusion task $\tau$. The top-ranked image is then selected as the preliminary pseudo ground truth $I_f^+$:
\begin{equation}
I_f^+ = \varphi(\mathcal{S}; \tau) = \arg\max_{I_f \in \mathcal{S}} \text{Score}(I_f;\tau).
\end{equation}

This selection mechanism ensures that the supervisory signal reflects task-specific preferences, while also mitigating bias from individual model bias. 

\paragraph{Fusion Refiner}
Although the preliminary pseudo ground truth $I_f^+$ is carefully selected, it may still suffer from issues such as local information loss, blurred structures, and subtle artifacts. To address this, we introduce a novel \textit{Fusion Refiner} $\mathcal{R}_\theta$, which adopts a divide-and-conquer strategy to improve fused image quality. The Fusion Refiner consists of three main units: a \textit{Decomposition Unit (DU)}, a \textit{Refinement Module (RM)}, and a \textit{Fusion Integrator (FI)}.

\textbf{Decomposition Unit.} Inspired by FusionBooster~\cite{Fusionbooster}, the DU comprises two lightweight CNN-based autoencoders. Each autoencoder decomposes the fused image $I_f^+$ into modality-specific components: $[I_{\text{part}A}, I_{\text{part}B}] = DU(I_f^+)$, where $I_{\text{part}A}$ approximates the latent content from modality $A$ (e.g., infrared), and $I_{\text{part}B}$ approximates that from modality $B$ (e.g., visible). This decomposition disentangles the mixed content of $I_f^+$ and facilitates modality-aware refinement.

\textbf{Refinement Module.} The RM is designed to restore fine details and preserve the overall fusion style of $I_f^+$ by leveraging both the decomposed components and the original source images. Since the decomposed components alone cannot fully recover lost content, we first extract high-frequency residuals using low-pass filtering with an average filter $\mathcal{H}$, yielding $I_{\text{part}A}^d = I_{\text{part}A} - I_{\text{part}A} * \mathcal{H}$ and $I_A^d = I_A - I_A * \mathcal{H}$. To effectively combine these complementary details, we compute an adaptive spatial weight map $W_A = \sigma(I_{\text{part}A}) \odot \varnothing_E(I_{\text{part}A})$, which guides the selective blending of high-frequency information from the source and decomposed domains. The fused detail layer is then obtained as $I_{\text{detail}}^A = W_A \odot I_A^d + (1 - W_A) \odot I_{\text{part}A}^d$. To avoid over-enhancement, a protection mask $P_A = \text{ReLU}(\varnothing_G(I_{\text{part}A}) - c)$ is applied, where $\varnothing_G(\cdot)$ denotes the gradient magnitude operator and $c$ is a fixed threshold. The final refined component is computed as $I_{\text{part}A}^+ = I_f^+ + \alpha_A \cdot I_{\text{detail}}^A \odot P_A$, where $\alpha_A$ is a learnable scalar parameter. The same procedure is applied symmetrically to $I_{\text{part}B}$ to obtain $I_{\text{part}B}^+$.

\textbf{Fusion Integrator.} FI is a lightweight dual-branch encoder-decoder network. Its architecture features two parallel CNN-based encoders that extract multi-scale features from the refined modality components ($I_{\text{part}A}^+$ and $I_{\text{part}B}^+$). The extracted features are then concatenated and processed by a decoder to synthesize the final image $I_f^{++}$.

In summary, our prior-guided pseudo ground-truth generation framework first selects a high-quality candidate fused image from multiple SOTA models and then refines it using a dedicated refinement network. By integrating task-specific prior knowledge with structural enhancement, the final pseudo ground truth $I_f^{++}$ provides a more reliable supervisory signal for training fusion models.

\subsection{Incremental Learning Strategy}
In image fusion, models are expected to accommodate diverse modalities and continuously adapt to new tasks in dynamic real-world scenarios. Additionally, data availability often varies significantly across tasks. To enable FusionFM to incrementally learn new tasks while effectively avoiding catastrophic forgetting, we adopt a dual-strategy approach that integrates EWC for parameter regularization and ER for data-level knowledge rehearsal.

\paragraph{Elastic Weight Consolidation.}
EWC introduces a regularization term to preserve important knowledge from previous tasks. After completing each task $k$, we compute the diagonal of the Fisher Information Matrix $F_k$ over model parameters $\theta$ as:
\begin{equation}
F_{k,i} = \mathbb{E}_{D_k} \left[ \left( \frac{\partial \log p(D_k|\theta_k^*)}{\partial \theta_i} \right)^2 \right],
\end{equation}
where $F_{k,i}$ approximates the importance of the $i$-th parameter $\theta_i$ for task $k$, and $D_k$ denotes the training dataset associated with task $k$. In practice, the log-likelihood term is typically approximated by $-\mathcal{L}(\theta_k^*)$, where $\mathcal{L}(\theta_k^*)$ is the task-specific loss evaluated at the optimal parameters. 

During training on the current task $T_{\text{current}}$, the EWC regularization term is computed as:
\begin{equation}
\mathcal{L}_{\text{EWC}} = \sum_{k < T_{\text{current}}} \sum_i \lambda F_{k,i} \left( \theta_i - \theta_{k,i}^* \right)^2,
\end{equation}
where $\lambda$ is a hyperparameter that controls the trade-off between stability and plasticity. $\theta_i$ denotes the current value of the $i$-th parameter, and $\theta_{k,i}^*$ is its optimal value from task $k$.

\paragraph{Experience Replay.}
While EWC provides parameter-level stability, it cannot fully address distributional shifts between tasks. We complement EWC with a theoretically motivated ER mechanism. For each completed task $k$, we maintain a representative subset $\mathcal{M}_k \subset D_k$ of size $m = \min(|D_k|, \texttt{memory\_size})$. This subset is created through random sampling to efficiently preserve the original data distribution of task $k$.
\begin{equation}
\mathcal{M}_k = \text{Sample}(D_k, m).
\end{equation}

The complete replay buffer aggregates memory samples from all previous tasks:
\begin{equation}
D_{\text{replay}} = \bigcup_{k < T_{\text{current}}} \mathcal{M}_k.
\end{equation}
This strategy provides an unbiased estimator of the joint distribution across previous tasks, enabling the model to rehearse past knowledge during new task learning. The exclusion of current task samples from the replay buffer prevents data leakage and ensures proper generalization.

\paragraph{Unified Training Objective.}
The complete training objective integrates two complementary components:
\begin{equation}
\mathcal{L}_{\text{unified}} = \mathcal{L}_{\text{FM}}(D_{\text{current}} \cup D_{\text{replay}}) + \lambda \mathcal{L}_{\text{EWC}},
\end{equation}
where $\mathcal{L}_{\text{FM}}(D_{\text{current}} \cup D_{\text{replay}})$ is the primary FM loss computed on the combined data from the current task ($D_{\text{current}}$) and replayed historical samples ($D_{\text{replay}}$), which collectively promote both adaptation to the current task and retention of prior knowledge. $\mathcal{L}_{\text{EWC}}$ is the EWC regularization term that preserves important parameters from previous tasks, weighted by the hyperparameter $\lambda$.
\begin{table*}[]
\caption{Quantitative comparison on fusion datasets. The best and second-best results are highlighted in \textbf{bold} and \underline{underline}.}
\resizebox{\textwidth}{!}{
\begin{tabular}{ccccccccccccccccccccc}
\toprule
\multirow{2}{*}{\textbf{}} & \multicolumn{6}{c}{\textbf{IVF-MSRS Dataset}}         & \multirow{2}{*}{\textbf{}} & \multicolumn{6}{c}{\textbf{IVF-LLVIP Dataset}}        & \multirow{2}{*}{\textbf{}} & \multicolumn{6}{c}{\textbf{IVF-FMB Dataset}}          \\ \cmidrule(lr){2-7} \cmidrule(lr){9-14} \cmidrule(lr){16-21}    
                           & EN   & SD    & SF    & AG   & VIF  & SCD  &                            & EN   & SD    & SF    & AG   & VIF  & SCD  &                            & EN   & SD    & SF    & AG   & VIF  & SCD  \\ \midrule
SwinFusion                 & 6.62 & 43.00 & 11.09 & 3.55 & 0.99 & 1.69 & SwinFusion                 & 7.42 & 53.07 & 16.36 & 4.88 & 0.88 & 1.58 & SwinFusion                 & 6.68 & 35.34 & 13.50 & 4.03 & 0.85 & 1.56 \\
SegMIF                     & 5.95 & 37.28 & 11.10 & 3.47 & 0.88 & 1.57 & SegMIF                     & 7.37 & 53.65 & 15.46 & 4.92 & 0.91 & 1.70 & SegMIF                     & 6.83 & \underline{40.76} & 13.63 & 4.14 & 0.81 & 1.72 \\
CDDFuse                    & 6.70 & 43.38 & 11.56 & 3.74 & 1.05 & 1.62 & CDDFuse                    & 7.38 & 51.15 & 16.42 & 4.49 & 0.87 & 1.58 & CDDFuse                    & \underline{6.93} & 39.16 & 13.83 & 4.13 & 0.84 & 1.67 \\
Diff-IF                    & 6.67 & 42.60 & 11.46 & 3.70 & 1.04 & 1.62 & Diff-IF                    & 7.45 & 51.41 & 16.82 & 4.52 & 0.91 & 1.47 & Diff-IF                    & 6.63 & 34.19 & 13.87 & 4.05 & 0.87 & 1.37 \\
EMMA                       & 6.71 & 44.13 & 11.56 & 3.76 & 0.97 & 1.63 & EMMA                       & 7.37 & 50.29 & 15.88 & 4.88 & 0.81 & 1.56 & EMMA                       & 6.77 & 36.80 & 15.00 & 4.67 & 0.83 & 1.50 \\
Text-IF                    & 6.73 & \underline{44.58} & 11.88 & \underline{3.87} & \underline{1.05} & 1.70 & Text-IF                    & 7.44 & \underline{54.35} & 19.66 & \underline{5.82} & \underline{0.94} & 1.60 & Text-IF                    & 6.74 & 34.66 & 15.05 & 4.50 & \textbf{0.95} & 1.71 \\
TC-MoA                     & 6.60 & 41.77 & 11.14 & 3.57 & 0.97 & 1.63 & TC-MoA                     & \underline{7.46} & 50.98 & 16.80 & 4.90 & 0.93 & 1.48 & TC-MoA                     & 6.62 & 32.42 & 13.51 & 3.96 & 0.86 & 1.43 \\
ReFusion                   & 6.69 & 42.47 & 11.58 & 3.80 & 1.00 & 1.67 & ReFusion                   & 7.44 & 51.40 & 17.04 & 4.93 & 0.88 & 1.55 & ReFusion                   & 6.79 & 36.67 & 15.17 & 4.54 & 0.87 & 1.62 \\
TDFusion                   & \underline{6.74} & 42.92 & 11.30 & 3.72 & 1.00 & \underline{1.79} & TDFusion                   & 7.38 & 51.98 & 16.92 & 4.93 & 0.91 & \underline{1.70} & TDFusion                   & 6.86 & 34.98 & 14.16 & 4.24 & 0.86 & \textbf{1.76} \\
GIFNet                     & 5.94 & 32.90 & \underline{12.72} & 3.37 & 0.58 & 1.41 & GIFNet                     & 7.01 & 46.70 & \textbf{21.67} & 5.77 & 0.64 & 1.51 & GIFNet                     & 6.90 & 39.11 & \underline{16.69} & \underline{5.03} & 0.59 & 1.73 \\
Our                        & \textbf{6.85} & \textbf{47.18} & \textbf{13.97} & \textbf{4.62} & \textbf{1.06} & \textbf{1.81} & Our                        & \textbf{7.62} & \textbf{62.77} & \underline{20.05} & \textbf{5.84} & \textbf{0.95} & \textbf{1.72} & Our                        & \textbf{7.09} & \textbf{47.62} & \textbf{17.59} & \textbf{5.25} & \underline{0.94} & \underline{1.73} \\ \midrule
\multirow{2}{*}{\textbf{}} & \multicolumn{6}{c}{\textbf{MIF-Harvard Medical dataset}}          & \multirow{2}{*}{\textbf{}} & \multicolumn{6}{c}{\textbf{MEF-SICE and MEFB Dataset}}          & \multirow{2}{*}{\textbf{}} & \multicolumn{6}{c}{\textbf{MFF-RealMFF and Lytro Dataset}}          \\ \cmidrule(lr){2-7} \cmidrule(lr){9-14} \cmidrule(lr){16-21}        
                           & EN   & SD    & SF    & AG   & VIF  & SCD  &                            & EN   & SD    & SF    & AG   & VIF  & Qabf &                            & EN   & SD    & SF    & AG   & VIF  & Qabf \\ \midrule
U2Fusion                   & 4.01 & 54.07 & 19.42 & 5.43 & 0.44 & 0.93 & U2Fusion                   & 6.39 & 36.37 & 10.64 & 3.18 & 1.16 & 0.55 & U2Fusion                   & 6.72 & 43.41 & 12.42 & 4.53 & 1.19 & 0.68 \\
SwinFusion                 & 4.17 & 73.44 & 21.85 & 6.07 & 0.64 & 1.58 & SwinFusion                 & 6.88 & 47.15 & \underline{19.07} & \underline{5.30} & 1.41 & 0.70 & SwinFusion                 & 7.09 & 51.58 & 13.73 & 4.79 & \underline{1.40} & 0.66 \\
CDDFuse                    & 4.29 & 72.41 & 23.72 & 6.22 & 0.59 & 1.54 & DeFusion                   & 6.68 & 46.18 & 12.87 & 3.68 & 1.02 & 0.57 & DeFusion                   & 7.07 & 51.08 & 10.54 & 3.85 & 1.26 & 0.65 \\
DDFM                       & 4.15 & 63.22 & 17.61 & 4.80 & 0.60 & 1.44 & HoLoCo                     & \underline{7.08} & 46.18 & 10.35 & 3.70 & 1.11 & 0.43 & ZMFF                       & 7.11 & 52.40 & 15.23 & 5.37 & 1.33 & 0.72 \\
Diff-IF                    & 4.15 & 72.33 & 27.38 & \underline{7.11} & \textbf{0.68} & 1.41 & MGDN                       & 6.93 & 44.90 & 14.97 & 4.61 & 1.33 & 0.64 & MGDN                       & 7.12 & 52.94 & 15.48 & 5.36 & 1.38 & 0.76 \\
EMMA                       & 4.17 & \underline{73.68} & 21.10 & 5.94 & 0.55 & \underline{1.60} & PSLPT                      & 6.67 & 38.17 & 12.14 & 3.17 & 0.35 & 0.18 & PSLPT                      & 7.11 & 52.50 & 13.66 & 4.88 & 0.84 & 0.47 \\
TC-MoA                     & 4.13 & 69.70  & 20.70 & 5.90 & 0.58 & 1.43 & TC-MoA                     & 7.05 & 44.49 & 16.42 & 4.71 & 1.13 & 0.69 & TC-MoA                     & 7.12 & 52.51 & 15.78 & \underline{5.39} & 1.35 & 0.74 \\
ReFusion                   & \textbf{5.11} & 70.59 & 25.22 & 7.11 & 0.65 & 1.51 & ReFusion                   & 6.84 & \underline{50.31} & 14.06 & 3.76 & 1.20 & 0.64 & ReFusion                   & 7.20 & 56.07 & 15.88 & 5.39 & 1.24 & 0.74 \\
LFDT                       & 4.23 & 71.23 & 25.48 & 6.66 & 0.64 & 1.34 & LFDT                       & 6.89 & 44.89 & 17.53 & 4.81 & \underline{1.42} & \textbf{0.77} & LFDT                       & 7.12 & 52.97 & 15.59 & 5.33 & 1.36 & \textbf{0.76} \\
GIFNet                     & 4.03 & 63.76 & \underline{27.06} & 7.02 & 0.41 & 1.18 & GIFNet                     & 7.08 & 44.34 & 17.81 & 4.65 & 1.12 & 0.44 & GIFNet                     & \underline{7.21} & \underline{63.61} & \underline{16.67} & 5.20 & 1.08 & 0.51 \\
Our                        & \underline{4.34} & \textbf{83.95} & \textbf{28.17} & \textbf{7.23} & \underline{0.66} & \textbf{1.73} & Our                        & \textbf{7.12} & \textbf{52.57} & \textbf{19.34} & \textbf{5.84} & \textbf{1.43} & \underline{0.72} & Our                        & \textbf{7.30} & \textbf{65.94} & \textbf{18.88} & \textbf{6.60} & \textbf{1.40} & \underline{0.71} \\ \bottomrule 
\end{tabular}
}
\label{Norm}
\end{table*}

\subsection{Training Strategy}
Training proceeds in two sequential stages to ensure stable convergence and effective task adaptation.
\paragraph{Pretraining} We first pretrain the DU by fine-tuning weights from FusionBooster on our dataset. The DU is optimized to reconstruct original modality images from the fused input $I_f^+$ under an $\ell_1$ reconstruction loss. At the same time, the FI is pretrained using the original source images $(I_A, I_B)$ as input. The FI is supervised with a hybrid objective that combines a well-established fusion loss (e.g., the SwinFusion loss, which integrates SSIM, texture, and intensity terms) with pseudo-label consistency against $I_f^+$. This stage equips the DU with reliable modality-specific decomposition ability and enables the FI to capture both general high-quality fusion characteristics and alignment with the selected ground truth.

\paragraph{Refinement Fine-tuning} After freezing the DU parameters to preserve stable decomposition, we fine-tune the FI as the core refinement module. The FI receives refined components \((I_{\text{part}A}^+, I_{\text{part}B}^+)\) generated by the non-trainable Refinement Module (RM). The training continues with the same hybrid loss function from Stage 1, encouraging the FI to produce a high-quality output \(I_f^{++}\) that maintains both fusion quality and pseudo-label consistency. While the RM itself contains no learnable layers, its task-adaptive scalar parameters \((\alpha_A, \alpha_B)\) are optimized jointly with the FI's weights during this stage.

This two-stage training strategy is designed to balance stability and adaptability. By pretraining the DU and FI separately, our model establishes a solid foundation in both modality decomposition and comprehensive fusion, which prevents unstable gradients and improves convergence. We then freeze the DU during fine-tuning to ensure stable decomposition, allowing the model to focus on refining fusion quality. The joint optimization of the FI and the RM’s scalar parameters enables adaptive detail enhancement without altering the core decomposition, effectively leveraging prior knowledge to meet task-specific requirements. This modular approach not only improves stability but also provides better control over detail preservation and artifact suppression in the final output.

\begin{figure*}[t]
\centering
\includegraphics[width=1\textwidth]{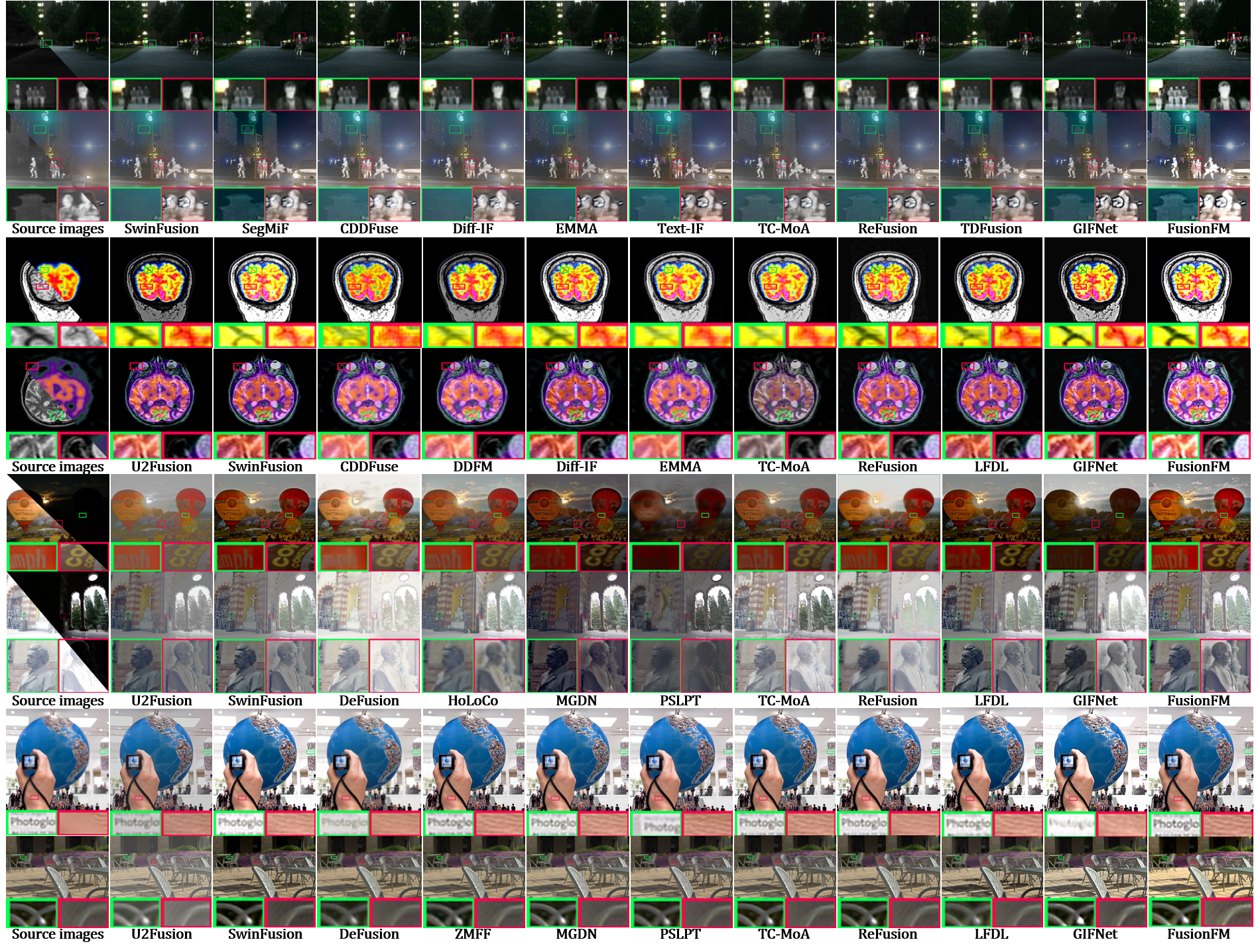  }
\caption{Qualitative results of previous methods and proposed method. From top to bottom, they are IVF, MIF, MEF, and MFF.}
\label{vis}
\end{figure*}

\begin{table}[t]
\centering
\caption{Task-specific weights assigned to each quality metric. A dash (--) indicates the metric is not used for that task.}
\label{tab:metric_weights}
\resizebox{0.9\columnwidth}{!}{
\begin{tabular}{lcccccc}
\toprule
\textbf{Task} & \textbf{EN} & \textbf{VIF} & \textbf{Qabf} & \textbf{SSIM} & \textbf{SD} & \textbf{SF} \\
\midrule
IVF, MIF      & 1           & 1            & 2             & 3             & --          & --          \\
MEF, MFF      & 1           & 5            & 6             & --            & 1           & 2           \\
\bottomrule
\end{tabular}
}
\end{table}

\section{Experiments}
\subsection{Experimental Setup and Evaluation}
\paragraph{Datasets and implementation details.} We evaluate our method on four representative image fusion tasks using the following datasets: (1) IVF: MSRS~\cite{PIAFusion}, FMB~\cite{SegM}, and LLVIP~\cite{LLIVP}; (2) MEF: SICE~\cite{SICE} and MEFB~\cite{MEFB}; (3) MFF: RealMFF~\cite{Real-MFF}, WFF-WHU~\cite{WFF-WHU}, and Lytro~\cite{Lytro}; and (4) MIF: Harvard Medical dataset~\cite{HarvardMedicalWebsite}. All experiments run on NVIDIA GeForce RTX 3090 GPUs using PyTorch. During training, images are randomly cropped to 128 × 128 pixels. Each task is trained for 25,000 iterations with batch size 32. The initial learning rate is set to 8e-4 and the regularization weight $\lambda$ is fixed at 1000.
\paragraph{Comparative methods and evaluation metrics} For IVF, we compare with U2Fusion, SwinFusion \cite{SwinFusion}, CDDFuse \cite{CDDFuse}, Diff-IF, EMMA \cite{EMMA}, Text-IF \cite{Text-if}, TC-MoA, ReFusion \cite{ReFusion}, TDFusion \cite{TDFusion}, and GIFNet. For MIF, we adopt SwinFusion, CDDFuse, EMMA, DDFM \cite{DDFM}, Diff-IF, LFDT \cite{LFDT}, ReFusion, GIFNet, and TC-MoA. For MEF, we consider SwinFusion, HoLoCo \cite{HoLoCo}, PSLPT \cite{PSLPT}, DeFusion \cite{DeFusion}, LFDT, ReFusion, TC-MoA, GIFNet, and MGDN \cite{MGDN}; and for MFF, the same set is used with ZMFF \cite{ZMFF} added. Additionally, we select six metrics to evaluate the fused result: EN, SD, SF, VIF, Qabf, SCD, and AG. Higher values for these metrics indicate better performance.
\paragraph{Teacher Model Pool} We construct a diverse pool of state-of-the-art fusion models to generate candidate fused images for pseudo ground-truth construction. For \textbf{IVF} and \textbf{MIF}, the pool includes SwinFusion~\cite{SwinFusion}, EMMA~\cite{EMMA}, ReFusion~\cite{ReFusion}, LFDT-Fusion~\cite{LFDT}, and GIFNet~\cite{GIFNet}, with Text-IF~\cite{Text-if} and TDFusion~\cite{TDFusion} additionally used for IVF. For \textbf{MEF} and \textbf{MFF}, we adopt SwinFusion, TC-MoA~\cite{TC-MoA}, LFDT-Fusion, and GIFNet, with HoLoCo~\cite{HoLoCo} (for MEF) and MGDN~\cite{MGDN} (for MFF) included respectively. This diverse pool ensures we generate a robust and comprehensive candidate set, which is crucial for effective pseudo-supervision.
\paragraph{Task-Specific Metric Weighting} Following Diff-IF~\cite{Diff-IF}, we assign task-dependent weights to reflect the distinct quality requirements of each fusion scenario. For IVF and MIF, where the goal is to preserve structural information and modality consistency, we emphasize metrics such as SSIM and Qabf. In contrast, MEF and MFF prioritize perceptual sharpness, exposure balance, and detail enhancement, thus placing more weight on VIF, SF, and Qabf. The weighting configuration is summarized in Table~\ref{tab:metric_weights}.

\paragraph{Weight Determination Strategy.}
For IVF and MIF, we adopt the heuristic weights from Diff-IF~\cite{Diff-IF}, which are empirically aligned with structural preservation goals in cross-modality fusion. For MEF and MFF, the weights are manually designed based on extensive visual inspection and task-specific fusion priorities. Although not derived from exhaustive optimization, this task-aware scheme has consistently provided stable and effective supervision for pseudo-label selection across diverse fusion settings.

\begin{figure*}[t]
\centering
\includegraphics[width=1\textwidth]{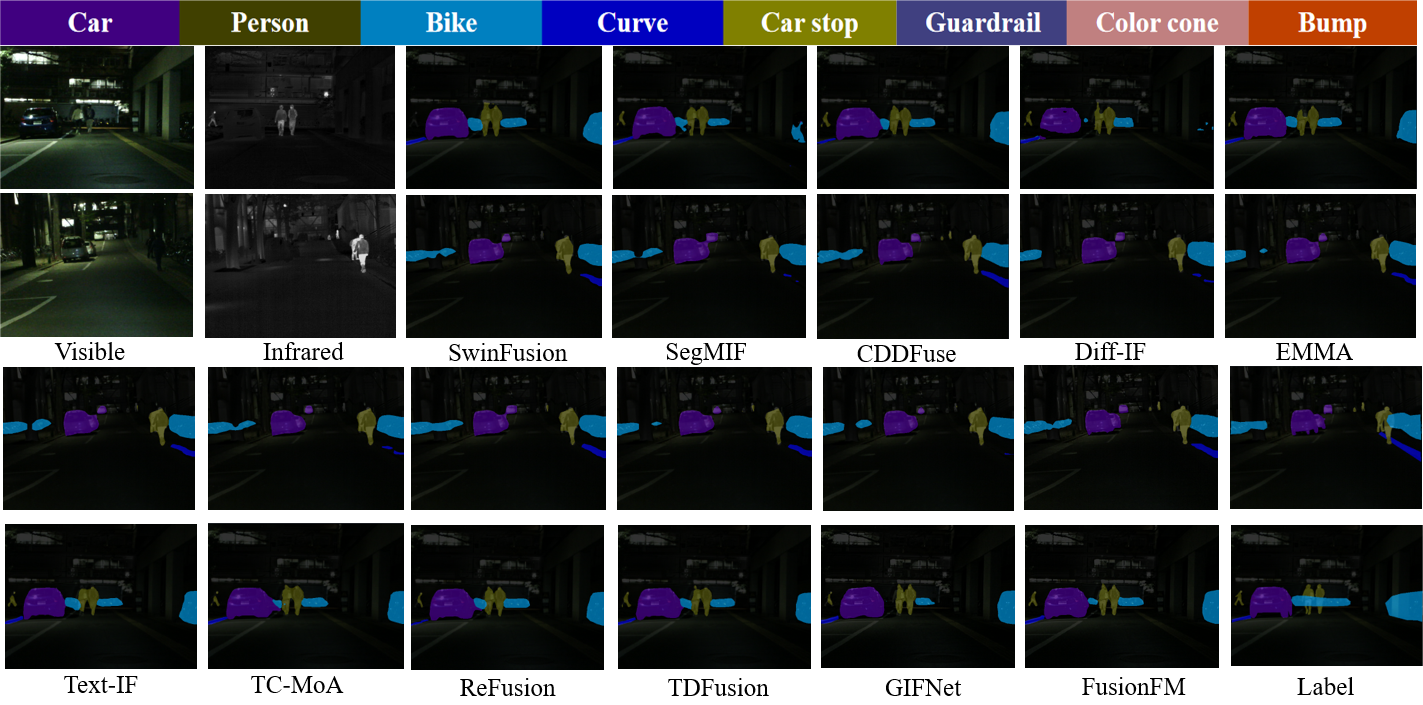}
\caption{Visualization results of semantic segmentation in two scenes on the MSRS dataset.}
\label{MSRS_Seg}
\end{figure*}

\begin{table*}[t]
\caption{Performance comparison of semantic segmentation.}
\resizebox{\textwidth}{!}{
\begin{tabular}{cccccccccccccccccccccc}
\toprule
\multirow{2}{*}{\textbf{}} & \multicolumn{10}{c}{\textbf{MSRS Dataset}}                                    & \multirow{2}{*}{\textbf{}} & \multicolumn{10}{c}{\textbf{FMB Dataset}}                                     \\ \cmidrule(lr){2-11} \cmidrule(lr){13-22}   
                           & Unl   & Car   & Per   & Bik   & Cur   & CS    & GD    & CC    & Bu    & mIoU  &                            & Roa   & Sid   & Bui   & Sig   & Veg   & Sky   & Per   & Car   & Pol   & mIoU  \\ \midrule
SwinFusion                 & 98.05 & \underline{87.21} & 65.92 & 64.01 & 54.69 & 65.60 & 54.32 & 53.55 & 62.26 & 67.46 & SwinFusion                 & 81.81 & 38.85 & 77.51 & 67.19 & 82.42 & 92.30 & 52.25 & 70.86 & 33.54 & 66.53 \\
SegMIF                     & 97.86 & 85.00 & 66.50  & 63.39 & 49.71 & 63.88 & 60.48 & 52.47 & 50.99 & 67.20 & SegMIF                     & 82.56 & 40.37 & 78.19 & \underline{67.62} & 82.62 & 92.31 & 52.43 & 71.17 & 33.39 & 66.74 \\
CDDFuse                    & \underline{98.15} & 87.00 & \underline{66.96} & 65.27 & 52.63 & 67.90 & 75.25 & 58.28 & 68.41 & 71.09 & CDDFuse                    & 83.20 & \underline{40.86} & 75.03 & 66.84 & 81.24 & 92.55 & 53.33 & 71.04 & 32.58 & 66.30 \\
Diff-IF                    & 98.05 & 86.44 & 65.99 & 66.74 & 54.11 & 68.12 & 72.99 & 57.18 & 65.20 & 70.53 & Diff-IF                    & 82.66 & 37.03 & 76.33 & 64.75 & 81.50 & 91.04 & 52.93 & \underline{71.67} & 33.55 & 65.72 \\
EMMA                       & 97.94 & 85.72 & 66.14 & 63.39 & 51.67 & 58.71 & \underline{76.15} & 55.00 & 65.68 & 68.16 & EMMA                       & 83.57 & 40.21 & \underline{79.05} & 65.80 & 82.34 & 92.30 & 49.88 & 70.91 & 32.19 & 66.47 \\
Text-IF                    & 98.12 & 87.12 & 66.07 & 66.50 & 55.07 & 70.02 & 72.45 & 58.28 & 66.32 & 71.22 & Text-IF                    & 81.71 & 39.46 & 77.67 & 65.61 & 81.96 & 91.66 & \underline{53.57} & 70.06 & \underline{34.10} & 66.20 \\
ReFusion                   & 98.14 & 87.03 & 66.91 & 66.53 & 55.44 & \underline{70.10} & 71.11 & 58.15 & 68.88 & \underline{71.48} & ReFusion                   & \underline{84.09} & 38.63 & 77.69 & 66.46 & \textbf{83.58} & 91.63 & 51.91 & 71.54 & 33.58 & \underline{66.57} \\
TC-MoA                     & 98.04 & 86.37 & 66.06 & 65.48 & 54.97 & 67.35 & 70.67 & 54.63 & 64.05 & 69.73 & TC-MoA                     & 81.42 & 38.72 & 77.75 & 66.09 & 81.37 & 91.85 & 52.08 & 69.19 & 33.04 & 65.72 \\
TDFusion                   & 98.13 & 86.92 & 66.53 & 66.12 & \underline{56.22} & 69.93 & 73.29 & \textbf{59.37} & 68.56 & 71.68 & TDFusion                   & 83.23 & 39.25 & 77.29 & 65.50 & 82.89 & \underline{92.40} & 49.84 & 70.87 & 33.29 & 66.06 \\
GIFNet                     & 98.00 & 86.84 & 65.10 & \underline{67.85} & 54.23 & 67.44 & \textbf{76.68} & 51.42 & 62.41 & 70.00 & GIFNet                     & 83.83 & 39.82 & 76.77 & 67.46 & 82.95 & 91.50 & 45.83 & 71.12 & 33.95 & 65.91 \\
Our                        & \textbf{98.32} & \textbf{88.27} & \textbf{68.40} & \textbf{68.61} & \textbf{60.54} & \textbf{73.06} & 74.87 & \underline{58.36} & \textbf{73.30} & \textbf{73.75} & Our                        & \textbf{85.57} & \textbf{41.20} & \textbf{79.77} & \textbf{67.89} & \underline{83.29} & \textbf{92.98} & \textbf{54.44} & \textbf{72.16} & \textbf{34.18} & \textbf{67.94} \\
\bottomrule
\end{tabular}
}
\label{se}
\end{table*}

\begin{figure*}[t]
\centering
\includegraphics[width=1\textwidth]{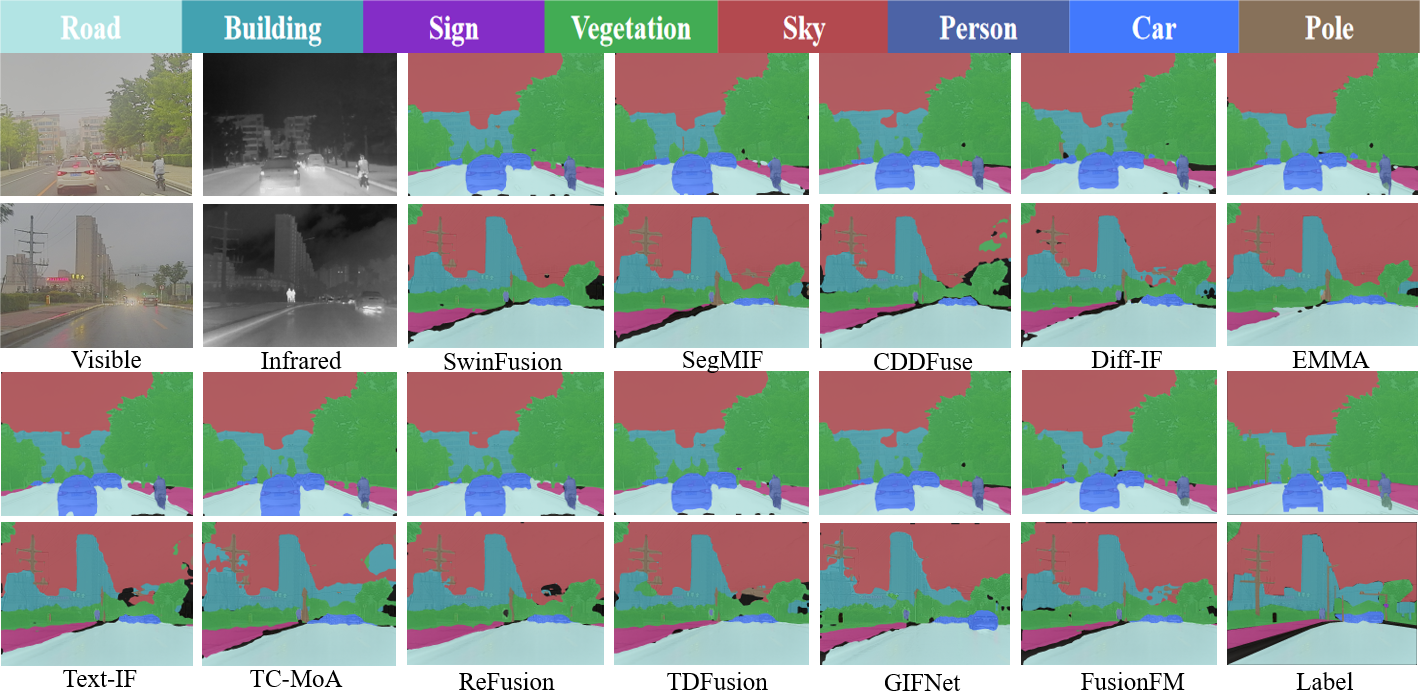}
\caption{Visualization results of semantic segmentation in two scenes on the FMB dataset.}
\label{FMB_Seg}
\end{figure*}

\subsection{Experimental Result}
\paragraph{Quantitative results} As shown in Table \ref{Norm}, our method consistently exhibits strong competitiveness across all datasets in the four fusion tasks. Specifically, in the IVF task, it frequently ranks first or second in metrics related to information content and structural texture preservation, while also performing well in terms of fidelity and contrast. Similarly, in the MIF task, although our approach slightly lags behind in the EN metric, it outperforms others in texture detail and information retention. For the MEF and MFF tasks, our model consistently achieves top-two performance across most evaluation metrics, demonstrating its ability to extract and integrate visual details effectively. These results confirm that FusionFM can produce high-quality fusion outputs and faithfully preserve scene content. Overall, the quantitative evidence clearly demonstrates the superior performance and strong generalization ability of our FusionFM.

\paragraph{Qualitative results} The visualization shown in Figure \ref{vis} demonstrates the significant advantages of FusionFM. As shown in the figure, our method can effectively preserve the texture details of the source image in the IVF scenario, especially on objects such as building textures and road signs. At the same time, it enhances the highlight information in the infrared image, captures the complex details of distant people, and reduces blur and artifacts. In the medical scenes in the third and fourth rows, FusionFM performs very well in preserving key tissue structures and functional information in medical images. In addition, in the MEF task involving exposure balance, our method significantly improves the brightness and contrast in extreme exposure regions and enhances color saturation. Compared with our method, other competitors often suffer from issues such as blurred edges and color distortion. Finally, in the MFF task, our method not only preserves the texture details in the transition areas between focused and unfocused regions but also ensures clear and consistent presentation of both the foreground and background. Overall, these results demonstrate the superior visual quality and strong generalizability of FusionFM.

\begin{figure*}[t]
\centering
\includegraphics[width=1\textwidth]{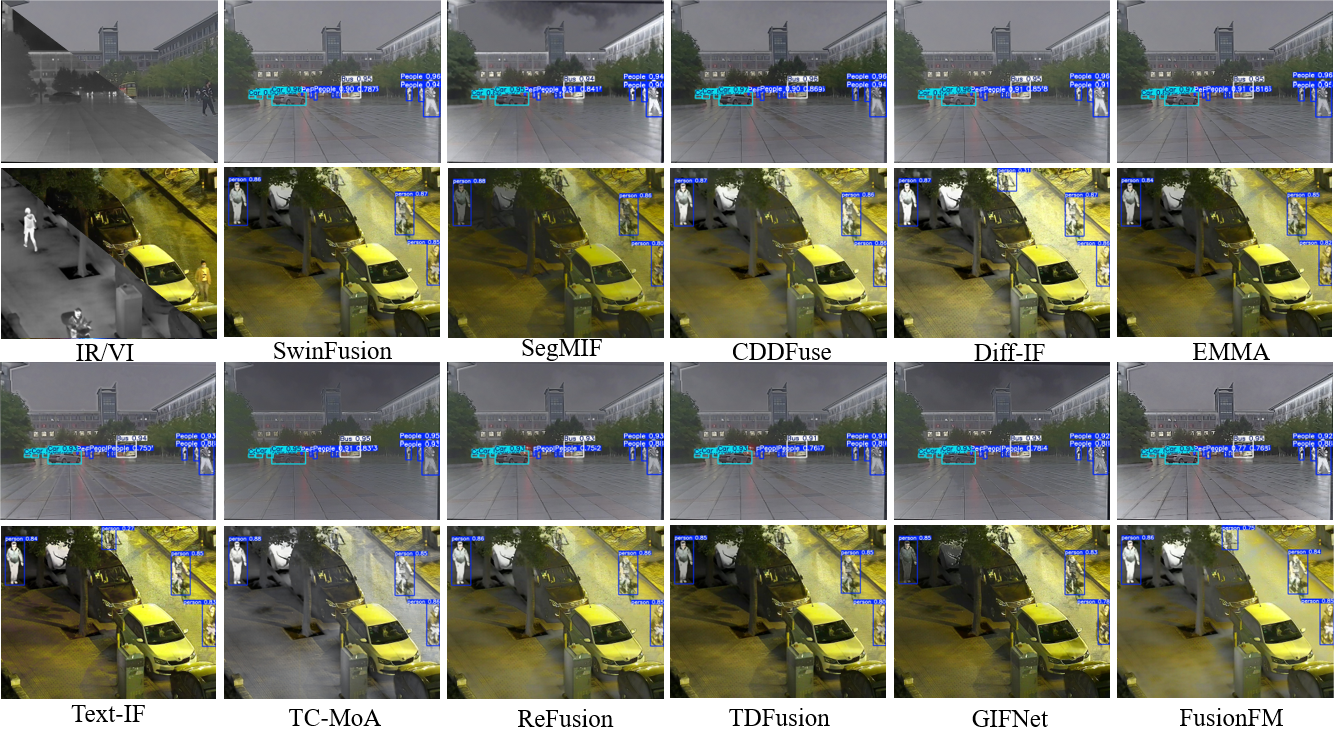}
\caption{Visualization results of target detection on MSRS and LLVIP datasets.}
\label{Object}
\end{figure*}

\begin{table}[t]
\centering
\caption{Performance comparison of object detection.}
\resizebox{\columnwidth}{!}{
\begin{tabular}{cccccccc}
\toprule
\multirow{2}{*}{\textbf{Method}} & \multicolumn{3}{c}{LLVIP-Dataset}           & \multirow{2}{*}{}    & \multicolumn{3}{c}{M3FD-Dataset}            \\ \cmidrule(lr){2-4} \cmidrule(lr){6-8}   
                                 & mAP50 & mAP75 & mAP50:95 &                      & mAP50 & mAP@75 & mAP@50:95 \\ \midrule
SwinFusion                       & 0.913    & 0.652    & 0.521         &                      & 0.891    & 0.673    & 0.622         \\
SegMIF                           & 0.932    & 0.664    & 0.601         &                      & 0.896    & 0.663    & \underline{0.624}         \\
CDDFuse                          & 0.942    & 0.682    & 0.614         &                      & 0.885    & 0.642    & 0.613         \\
Diff-IF                          & \underline{0.949}    & 0.693    & 0.637         &                      & 0.888    & 0.651    & 0.610         \\
EMMA                             & 0.935    & 0.652    & 0.594         &                      & 0.883    & 0.642    & 0.607         \\
Text-IF                          & 0.940    & 0.695    & 0.620         &                      & 0.897    & 0.664    & 0.623         \\
TC-MoA                           & 0.920    & 0.690    & 0.631         &                      & 0.887    & 0.659    & 0.604         \\
ReFusion                         & 0.942    & 0.701    & 0.639         &                      & 0.892    & 0.671    & 0.614         \\
TDFusion                         & 0.945    & \underline{0.712}    & \underline{0.642}         &                      & \underline{0.898}    & \underline{0.690}    & 0.623         \\
GIFNet                           & 0.926    & 0.687    & 0.632         &                      & 0.892    & 0.662    & 0.606         \\
Our                              & \textbf{0.968}    & \textbf{0.762}    & \textbf{0.660}         & \multicolumn{1}{l}{} & \textbf{0.907}    & \textbf{0.696}    & \textbf{0.633}        \\ \bottomrule
\end{tabular}
}
\label{ob}
\end{table}

\subsection{Downstream applications.} To further evaluate the actual advantages of the fusion model, we conducted experiments on two representative downstream tasks. 

\paragraph{Semantic Segmentation} 
We adopted SegFormer \cite{SegFormer} as the base model and retrained it under identical experimental settings using different fused images as inputs. Table~\ref{se} reports the results on the MSRS and FMB datasets. Our method consistently outperforms other approaches across most categories and achieves the highest mIoU on both datasets. 

Specifically, on the MSRS dataset, FusionFM achieves the best performance in almost all categories, such as \textit{Car}, \textit{Person}, and \textit{Building}, and attains a remarkable overall mIoU of 73.75\%, surpassing the second-best method by more than 2 points. This suggests that our fusion strategy better preserves fine-grained object boundaries and maintains semantic consistency. On the FMB dataset, FusionFM also achieves leading performance in most classes, including challenging categories such as \textit{Sign}, \textit{Person}, and \textit{Car}, and reaches the highest mIoU of 67.94\%. The consistent gains across these two benchmarks demonstrate that our model effectively provides segmentation networks with richer structural details and more discriminative fused features, thereby enhancing pixel-level scene understanding in both urban and natural environments.

To further illustrate these improvements, we present qualitative results in Fig.~\ref{MSRS_Seg} and Fig.~\ref{FMB_Seg}. On the MSRS dataset, FusionFM successfully segments small background pedestrians as well as complete primary targets, while other methods such as SwinFusion and EMMA sometimes fail to capture the targets fully, leading to noticeable segmentation omissions. On the more challenging FMB dataset, FusionFM continues to show strong robustness: it more clearly distinguishes roads from sidewalks and provides accurate segmentation of vehicles, pedestrians, and traffic signs under varying illumination, occlusions, and large scale differences. By contrast, competing methods may occasionally suffer from blurred boundaries or category confusions. Overall, these results confirm that FusionFM delivers more reliable pixel-level semantic information in complex urban scenes, offering stronger support for downstream tasks.

\begin{table}[t]
\caption{Quantitative results of the ablation studies.}
\resizebox{\columnwidth}{!}{
\begin{tabular}{lcccclcccc}
\toprule
\multirow{2}{*}{\textbf{}} & \multicolumn{4}{c}{IVF-MSRS Dataset} & \multicolumn{1}{c}{\multirow{2}{*}{}} & \multicolumn{4}{c}{MEF-SICE and MEFB   Dataset} \\ \cmidrule(lr){2-5} \cmidrule(lr){7-10}
                           & EN      & SD       & VIF    & SCD    & \multicolumn{1}{c}{}                  & EN         & SD         & VIF       & Qabf      \\ \midrule
Exp.I                      & 6.24    & 30.53    & 0.78   & 1.48   &                                       & 6.23       & 51.89      & 1.05      & 0.59      \\
Exp.II                     & 6.67    & 42.56    & 1.01   & 1.65   &                                       & 6.77       & 46.23      & 1.37      & 0.69      \\
Exp.III                    & 6.79    & 44.47    & 1.02   & 1.70   &                                       & 7.06       & 50.19      & 1.25      & 0.56      \\
Exp.IV                     & 6.80    & 45.47    & 1.03   & 1.75   &                                       & 7.05       & 48.75      & 1.34      & 0.65      \\
Exp.V                      & 6.70    & 42.44    & 0.99   & 1.65   &                                       & 7.00       & 47.41      & 1.29      & 0.64      \\
Exp.VI                     & 6.66    & 40.94    & 0.95   & 1.52   &                                       & 6.91       & 48.14      & 1.30      & 0.69      \\
Exp.VII                    & 6.40    & 35.53    & 0.88   & 1.48   &                                       & 6.84       & 44.31      & 1.28      & 0.62      \\
Our                        & \textbf{6.85}    & \textbf{47.18}    & \textbf{1.06}   & \textbf{1.81}   &                                       & \textbf{7.12}       & \textbf{52.57}      & \textbf{1.43}      & \textbf{0.72}     \\ \bottomrule
\end{tabular}
}
\label{Ablation}
\end{table}

\paragraph{Object Detection} For detection, we employed YOLOv11 \cite{yolov11} as the baseline model and retrained it with various fused images. As shown in Table~\ref{ob}, FusionFM achieves the best detection performance on both the LLVIP and M3FD datasets \cite{TarDAL}. 

On the LLVIP dataset, FusionFM reaches an mAP50 of 0.968 and an mAP@75 of 0.762, which exceed the second-best results by significant margins (over 1.9\% and 5.0\%, respectively). The overall mAP50:95 also improves to 0.660, confirming that our method benefits detection not only at loose thresholds but also under stricter localization requirements. On the M3FD dataset, FusionFM consistently delivers the best performance, with 0.907 mAP50, 0.696 mAP@75, and 0.633 mAP50:95. Compared with competitive baselines such as TDFusion and ReFusion, our method shows consistent improvements, demonstrating its robustness across diverse scenarios.

To better understand these improvements, we also present qualitative detection results, as illustrated in Fig.~\ref{Object}. FusionFM enables more precise detection of objects with high confidence, even under challenging conditions. For example, in the LLVIP scenes, our method successfully detects a small pedestrian riding a bicycle in the upper-right corner, whereas methods such as GIFNet and TDFusion fail to capture this target. These visual comparisons confirm that FusionFM enhances object-level cues—such as edges, textures, and thermal information—thereby supporting more accurate localization and recognition in complex or low-visibility environments. The consistent superiority across both quantitative and qualitative evaluations highlights the generalizability and robustness of our fusion strategy for downstream detection tasks.

\begin{figure}[t]
\centering
\includegraphics[width=\columnwidth]{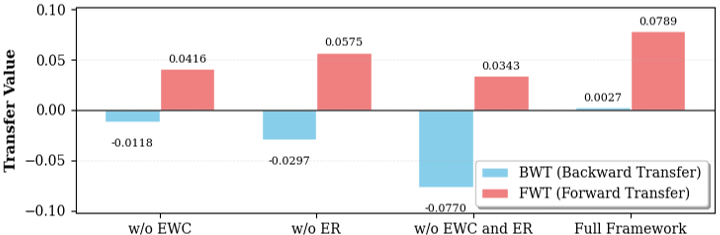}
\caption{Ablation Study: BWT and FWT Performance.}
\label{a1}
\end{figure}
\subsection{Ablation Studies}
\paragraph{Source-Fusion Image Coupling} We compare \textit{FusionFM} to a simple flow matching baseline. While simple flow matching also uses an optimal transfer-based objective to regress the vector field, it starts with Gaussian noise, i.e., $\mathbf{p(x_0)} \sim \mathcal{N}(0, I)$. The results (\textbf{Exp.I}) in Table \ref{Ablation} show that starting directly from the source image significantly improves the quality of the fused image.
\paragraph{Pseudo-truth Generation} We investigate the impact of fusion priors on model performance. First, we remove the selection strategy $\varphi$ and instead use a single fusion model (SwinFusion) as the prior (\textbf{Exp.II}). As shown in Table~\ref{Ablation}, this leads to a notable performance drop, highlighting the importance of selecting high-quality priors. Second, we remove the \textit{Fusion Refiner} (\textbf{Exp. III}), which also results in degraded fusion quality, indicating the necessity of refining pseudo labels. Finally, we replace our \textit{Fusion Refiner} (\textbf{Exp. IV}) with the \textit{Fusion Booster} proposed by Cheng~\cite{Fusionbooster}. The results further confirm the effectiveness of our selection-and-refinement strategy in generating reliable ground truths.
\paragraph{Incremental Learning Strategy} We conduct ablation studies to assess the contributions of key components in our continual learning framework. To clarify, BWT (Backward Transfer) measures the retention of knowledge on previous tasks, while FWT (Forward Transfer) evaluates how prior knowledge facilitates learning new tasks. As shown in Figure~\ref{a1}, removing the EWC regularization (\textbf{Exp.V}) leads to a noticeable drop in BWT, indicating increased forgetting on earlier tasks. Disabling the experience replay buffer (\textbf{Exp.VI}) also degrades stability and overall performance. When both mechanisms are removed (\textbf{Exp.VII}), performance declines further across all metrics, underscoring the complementary roles of parameter-level regularization and data-level rehearsal. In contrast, our complete framework achieves superior results on BWT, FWT, and various fusion metrics, demonstrating a balanced trade-off between plasticity and stability. These findings are further substantiated by the quantitative comparisons in Table~\ref{Ablation}.

\begin{table}[]
\caption{Computational efficiency comparison in terms of inference time and model size.}
\resizebox{\columnwidth}{!}{
\begin{tabular}{lccccccc}
\toprule
\textbf{Model}  & \textbf{Text-IF} & \textbf{TC-MoA} & \textbf{GIFNet} & \textbf{LFDT} & \textbf{DDFM} & \textbf{Diff-IF} & \textbf{Ours} \\ \midrule
Time (640$\times$480) [s]   & 0.560 & 0.613 & \underline{0.247} & 0.264 & 59.601 & 1.631 & \textbf{0.008} \\
Time (1280$\times$1024) [s] & 1.794 & 1.477 & 0.940 & \underline{0.873} & Out & 7.836 & \textbf{0.191} \\
Params (M)                  & 215.117 & 340.887 & \underline{3.291} & 20.280 & 552.814 & 23.736 & \textbf{3.228} \\ 
\bottomrule
\end{tabular}
}
\label{c}
\end{table}

\subsection{Computational Efficiency} 
FusionFM achieves high computational efficiency by adopting a flow matching–based one-shot sampling strategy, which eliminates the need for the costly multi-step iterative denoising process commonly required in diffusion-based fusion models. Instead of performing dozens or even hundreds of refinement steps, our method generates the fused result in a single forward pass, substantially reducing inference latency. As reported in Table~\ref{c}, FusionFM attains the fastest runtime under both standard-resolution (640$\times$480) and high-resolution (1280$\times$1024) inputs, while maintaining a lightweight U-Net backbone with only 3.2M parameters. These properties make FusionFM highly suitable for real-time applications and deployment in resource-constrained environments, where both efficiency and accuracy are critical.

\section{Conclusion}
We propose FusionFM, a flow matching framework for generic multi-modal image fusion, designed to enhance sampling efficiency, fusion quality, and generalization. First, we improve sampling efficiency by directly learning transport paths between source images and fused outputs, making the model faster and more stable than diffusion-based approaches. Second, we introduce a prior-guided pseudo-labeling strategy that selects optimal fusion priors from multiple SOTA models and refines them through a dedicated Fusion Refiner. Third, we enhance continual learning by integrating elastic weight consolidation and experience replay, ensuring stable performance across evolving tasks. Additionally, FusionFM is lightweight thanks to the high structural consistency between source and fused images.

\section*{Acknowledgments}
This research was supported by the National Natural Science Foundation of China under Grant Nos. 62202362 and 62302073, by the China Postdoctoral Science Foundation under Grant Nos. 2022TQ0247 and 2023M742742, by the Guangdong Basic and Applied Basic Research Foundation under Grant Nos. 2024A1515011626 and 2025A1515012949, and by the Science and Technology Projects in Guangzhou under Grant No. 2023A04J0397.

\bibliographystyle{IEEEtran}   % 参考文献样式（IEEE 格式）
\bibliography{FusionFM_bib}    % 你的 .bib 文件名（不要加后缀 .bib）

\newpage

 
\vspace{11pt}

% \end{IEEEbiography}

\vspace{11pt}


\vfill

\end{document}